\documentclass[journal]{IEEEtran}
\usepackage[noadjust]{cite}

\usepackage{amsmath}
\usepackage{amssymb}
\usepackage[mathscr]{euscript}
\usepackage[utf8]{inputenc}
\usepackage{flafter}
\newcommand{\norm}[1]{\left\lVert#1\right\rVert}

\newcommand{\mn}{MobileNet-V2}
\newcommand{\rn}{ResNet-50}
\newcommand{\nx}{Xavier}
\newcommand{\ignore}[1]{}
\usepackage{comment}

\usepackage{pgfplots}
\usepgfplotslibrary{clickable}

\usepackage{color}
\newcommand{\blue}[1]{\textcolor{blue}{#1}} 
\definecolor{ao}{rgb}{0.0, 0.5, 0.0}
\definecolor{amber}{rgb}{1.0, 0.60, 0.0}

\usepackage{algorithm,algorithmicx,algpseudocode}

\usepackage{array}
\usepackage{makecell}
\newcolumntype{C}[1]{>{\centering\arraybackslash}p{#1}}

\usepackage{scalerel,stackengine}
\stackMath
\newcommand\reallywidehat[1]{%
\savestack{\tmpbox}{\stretchto{%
  \scaleto{%
    \scalerel*[\widthof{\ensuremath{#1}}]{\kern.1pt\mathchar"0362\kern.1pt}%
    {\rule{0ex}{\textheight}}
  }{\textheight}%
}{2.4ex}}%
\stackon[-6.9pt]{#1}{\tmpbox}%
}
\newcommand{\figref}[1]{Fig.~\ref{#1}}
\newcommand{\secref}[1]{Section~\ref{#1}}
\newcommand{\tabref}[1]{Table~\ref{#1}}

\newcommand{\algoref}[1]{Algorithm~\ref{#1}}

\usepackage{hyperref}
\hypersetup{pdfborderstyle={/S/U/W 1}}

\usepackage{graphicx}
\usepackage{subcaption}

\usepackage{color}

\makeatletter
\newcommand*\bigcdot{\mathpalette\bigcdot@{.5}}
\newcommand*\bigcdot@[2]{\mathbin{\vcenter{\hbox{\scalebox{#2}{$\m@th#1\bullet$}}}}}
\makeatother

\usepackage{tikz}
\usetikzlibrary{matrix,calc}
\usetikzlibrary{snakes}
\pgfplotsset{compat=1.12}

\makeatletter
\newcommand{\printfnsymbol}[1]{%
  \textsuperscript{\@fnsymbol{#1}}%
}

\makeatother

\begin{document}
	
	
%

\title{REVAMP\textsuperscript{2}T: Real-time Edge Video Analytics for Multi-camera Privacy-aware Pedestrian Tracking\\}


%
%
%

\author{Christopher~Neff\IEEEauthorrefmark{1},~\IEEEmembership{Student Member,~IEEE,}
        Mat\'{i}as~Mendieta\IEEEauthorrefmark{1},~\IEEEmembership{Student Member,~IEEE,}
        Shrey~Mohan,~\IEEEmembership{Student Member,~IEEE,}
        Mohammadreza~Baharani,~\IEEEmembership{Student Member,~IEEE,}
        Samuel Rogers,~\IEEEmembership{Student Member,~IEEE,}
		Hamed~Tabkhi,~\IEEEmembership{Member,~IEEE}
\thanks{The authors are with the Electrical and Computer Engineering Department, The University of North Carolina at Charlotte, Charlotte,
	NC, 28223 USA  (e-mail: cneff1@uncc.edu, mmendiet@uncc.edu, smohan7@uncc.edu, mbaharan@uncc.edu,  sroger48@uncc.edu, htabkhiv@uncc.edu).
	
	\textsuperscript{*} Corresponding authors have equal contribution.
	
\textsuperscript{\textcopyright} 2019 IEEE.  Personal use of this material is permitted.  Permission from IEEE must be obtained for all other uses, in any current or future media, including reprinting/republishing this material for advertising or promotional purposes, creating new collective works, for resale or redistribution to servers or lists, or reuse of any copyrighted component of this work in other works. DOI: \href{https://doi.org/10.1109/JIOT.2019.2954804}{https://doi.org/10.1109/JIOT.2019.2954804}}}

%
%

\markboth{Published as an article paper at IEEE Internet of Things Journal: Privacy and Security in Distributed Edge Computing and Evolving IoT}%
{Neff \MakeLowercase{\textit{et al.}}: REVAMP\textsuperscript{2}T: Real-time Edge Video Analytics for Multi-camera Privacy-aware Pedestrian Tracking}
%


\maketitle

\begin{abstract}
This article presents REVAMP\textsuperscript{2}T, Real-time Edge Video Analytics for Multi-camera Privacy-aware Pedestrian Tracking, as an integrated end-to-end IoT system for privacy-built-in decentralized situational awareness. REVAMP\textsuperscript{2}T presents novel algorithmic and system constructs to push deep learning and video analytics next to IoT devices (i.e. video cameras). On the algorithm side, REVAMP\textsuperscript{2}T proposes a unified integrated computer vision pipeline for detection, re-identification, and tracking across multiple cameras without the need for storing the streaming data. At the same time, it avoids facial recognition, and tracks and re-identifies pedestrians based on their key features at runtime. On the IoT system side, REVAMP\textsuperscript{2}T provides infrastructure to maximize hardware utilization on the edge, orchestrates global communications, and provides system-wide re-identification, without the use of personally identifiable information, for a distributed IoT network. For the results and evaluation, this article also proposes a new metric, Accuracy$\bigcdot$Efficiency (\AE), for holistic evaluation of IoT systems for real-time video analytics based on accuracy, performance, and power efficiency. REVAMP\textsuperscript{2}T outperforms current state-of-the-art by as much as thirteen-fold \AE~improvement.
\end{abstract}

\begin{IEEEkeywords}
Edge Computing, Video Analytics, Deep Learning, Re-identification, Pedestrian Tracking, Privacy.
\end{IEEEkeywords}
\vspace{-10pt}

\IEEEpeerreviewmaketitle

\section{Introduction}

The emerging wave of Internet of Things (IoTs), scarcity of bandwidth resources, and tight latency awareness is pushing system designers to extend cloud computing to the edge of the network. Edge computing (and also fog computing \cite{Fog-edge}) refers to a group of technologies allowing cooperative computation at the edge of the network \cite{MEC-vision,MEC-city}. Ambient computer vision and real-time video analytics are the major classes of applications that requires edge computing for human-like vision processing over a large geographic area \cite{MEC-city,IEEE-mag-MEC,fog-princeton,Cloudlets}.


\begin{figure}[t] 
	\centering
	\includegraphics[width=1\linewidth, trim= 0 0 0 0,clip]{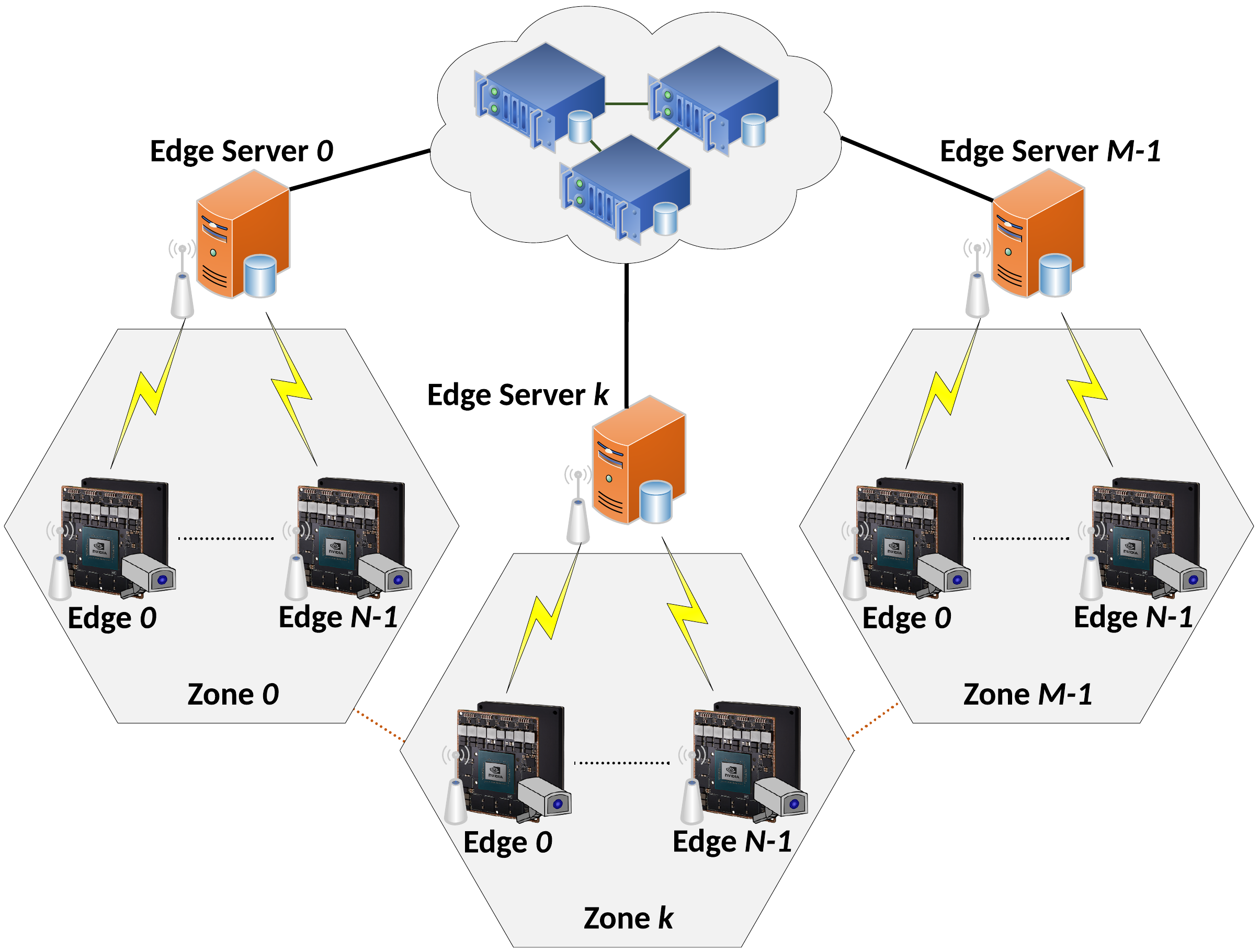}
	\vspace{-10pt}
	\captionsetup{justification=centering}
	\caption{Hierarchical System Overview}
	\vspace{-15pt}
	\label{fig:overview}
\end{figure}

Recent advances in machine learning, particularly deep learning, have driven the development of more advanced video analytics and surveillance technologies. This includes everything from simple license plate scanners that search for stolen vehicles, to facial recognition and pedestrian tracking. These applications often rely on a cloud computing paradigm for mass video collection and processing on a centralized computing server. The cloud computing paradigm introduces significant technical and social/ethical concerns for such applications. On the technical side, cloud computing leads to mass recording and storage of raw video data, which result in significant costs and limits scalability. At the same time, cloud computing is not applicable to many inherently real-time and time-sensitive video analytics.

On the social perspective, the broad net cast by typical surveillance approaches means that large amounts of personal information are incidentally collected and stored. This has led to significant push-back by privacy advocates against any expansions to video surveillance systems As an example, multiple cities in the US have imposed bans on all deployment of facial recognition and tracking technologies \cite{NyTimes}. European Union regulators are also considering new restrictions on AI-driven surveillance \cite{EU}. To address both technical and ethical concerns, novel approaches are required to address both IoT systems design and privacy challenges in a holistic manner across entire computing stack from algorithm design to computation mapping, communication, and system-level synchronization.



This paper introduces novel Real-time Edge Video Analytics for Multi-camera Privacy-aware Pedestrian Tracking, or REVAMP\textsuperscript{2}T. REVAMP\textsuperscript{2}T is able to track pedestrians across multiple cameras without ever transferring raw video or other forms of personally identifiable information. Fig.\ref{fig:overview} presents our proposed REVAMP\textsuperscript{2}T IoT system. Each IoT device (edge nodes) contains cameras equipped with the NVIDIA AGX Xavier \cite{Xavier} embedded platform, running a deep learning based video analytics pipeline, for real-time pedestrian detection and tracking over streaming pixels. In keeping with the concept of the "right to be forgotten" that was recently enshrined in EU law, our system does not rely on a static identity database. Instead, unique identities are generated when pedestrians first enter the view of a camera in our system and forgotten when those individuals are no longer being actively tracked by any part of our system.

Overall, REVAMP\textsuperscript{2}T achieves a pedestrian re-identification accuracy of 74.8\% (only 4.3\% below the current state-of-the-art \cite{DeepCC}) on the DukeMTMC dataset \cite{ristani2016MTMC}, while achieving more than two times the real-time FPS and consuming 1/5th of the power compared to \cite{DeepCC}. A balance was struck between algorithmic Accuracy and system Efficiency, measured by Accuracy$\bigcdot$Efficiency (\AE). Our system has high scalability potential in a multi-camera IoT environment while never sacrificing personal privacy.

The rest of this article is organized as follows: \secref{sec:Related} briefly reviews the other works related to the various components of our system. \secref{sec:Privacy} describes the privacy threat models. Next, \secref{Algorithm} details the design and implementation of our edge tracking algorithms. In \secref{System}, we describe our edge-to-network infrastructure for enabling multi-camera identity tracking. Evaluation of our complete system is detailed in \secref{Results}. Finally in \secref{Conclusion} we discuss our final conclusions as well as areas for future work.

\vspace{-5pt}
\section{Related Work}\label{sec:Related}
\vspace{-5pt}

\subsection{Pedestrian Detection, Re-Identification, and Tracking}
With the rapid advancements made in deep learning, a plethora of work has been published on pedestrian detection. Such models include region proposal networks like Faster-RCNN \cite{RCNN}, single shot detectors like SSD \cite{SSD} and YOLO \cite{yolov3}, as well as pose-estimation models like DeeperCut \cite{deepcut} and OpenPose \cite{openpose}. When analyzing these algorithms in light of edge-capable real-time performance, MobileNet-SSD, TinyYOLOv3, and OpenPose show promising results. 

The heart of pedestrian tracking is consistent re-identification (ReID) of those pedestrians throughout the frames of videos across multiple cameras. Similarly, on the re-identification side, recent methods leverage CNNs to extract unique features among persons \cite{zhang2017alignedreid,ristani2018features,zhou2019osnet, Shen_2018_ECCV,Li,Unsupervised_LI, Zhu, Xiao}. The work in \cite{Zhang} learns the spatial and temporal behavior of objects by translating the feature map of the Region of Interest (RoI) into an adaptive body-action unit. \cite{Dai} uses bidirectional Long-Short-Term-Memory (LSTM) neural networks to learn the spatial and temporal behavior of people throughout the video. Triplet loss \cite{DBLP:journals/corr/abs-1812-06576,Weinberger:2009:DML:1577069.1577078,Hermans2017InDO} is another promising technique to train the network with the goal of clustering classes in a way that IDs with the same type have minimum distance among each other, while examples from different categories are separated by a large margin. 

Pedestrian tracking systems often rely on prediction models to create insight on the changes in movement over time and empowers object re-identification. Object tracking has been tried using spatial masking and Kalman filter techniques for single and multiple object tracking \cite{track_kernal,kalman_adapt,kalman_multi}. In contrast, there is an interest in leveraging LSTM networks for prediction and tracking. One pronounced example is ROLO \cite{LSTM_yolo}, which uses YOLOv1 as its feature extractor, combined with LSTMs. Similarly, \cite{LSTM_vgg} uses VGG-16 for feature extraction and inputs the 500x1 feature vector into an LSTM. LSTM networks have been shown to provide lower Mean Squared Error in single object and fewer ID switches in multi-object tasks. However, the approaches in \cite{LSTM_vgg} and \cite{LSTM_yolo} often show very low accuracy as they are not customized for human objects.

Overall, the current state of pedestrian tracking algorithms struggle with limited focus and lack of privacy preservation. First, they look at the problem of pedestrian tracking in isolation, whether solely by detection, only re-identification using image crops, or just tracking with trajectories. However, these approaches do not analyze the problem in a holistic manner, which would require designing a pipeline to understand, integrate, and correlate these three functions into a single intelligent system. Second, the idea of privacy preservation and online functionality are lost with this narrowly focused approach. The previous works typically rely on the storage of large time segments of video data or image crops, degrading privacy preservation. Similarly, many works propose facial recognition techniques \cite{face-rgbd,facenet,face-gmm, face-eccv16, sighthound}, which also gravely compromises the privacy of tracked persons, requiring the pre-loaded and long-term storage of personally identifiable information like a facial database. At the same time, existing approaches typically analyzes the data offline with the ability to move forward and backward in time to maximize their algorithm accuracy scores, making edge deployable operation of these approaches impractical. In contrast to existing work, this article proposes a shift to non-personal and data private pedestrian tracking, improving upon our previous work in re-identification \cite{Baharani} and LSTM tracking \cite{Keytrack} for a holistic algorithm pipeline and fully edge capable design.

\vspace{-10pt}
\subsection{IoT Systems for Edge Video Analytics}

The concept and motivation behind edge computing has been described in a number of recent publications \cite{fog-princeton,Cloudlets,SWARM,MEC-city,IEEE-mag-MEC,IoT-app,Edge-vision,Fog-edge,Fog-computing}. However, there are very few works that present a distributed IoT system for video analytics and real-time tracking. \cite{vehicleMTMC} proposes a basic system framework for vehicle detection and tracking across multiple cameras. The approach uses positional matching for re-identification, relying on GPS coordinates, known distances, and time synchronization between cameras. The Gabriel project from CMU \cite{Ha2014Gabriel} is a wearable cognitive assistance system where the images captured by a mobile device are processed by the edge node to analyze what the user is seeing, and provide the user with cues as to what is in the scene (for example, recognizing a person). In the VisFlow project from Microsoft, Lu et al. \cite{Lu2016visflow} describe a system that can analyze feeds from multiple cameras. In particular, they describe a dataflow platform for vision queries that is built on top of the SCOPE dataflow engine, that offers general SQL syntax, and supports added user-defined operators such as extractors, processors, reducers, and combiners. \cite{BIPCC} proposed a method for single camera multi-target tracking in terms of the Binary Integer Program, and can incur online, real-time results on hardware. However, the system does not scale to multi-camera systems. The current state-of-the-art in Multi-Target Multi-Camera (MTMC) systems is DeepCC \cite{DeepCC}. This approach uses OpenPose for detections, a deep learning triplet loss ReID network for visual data association, and trajectory tracklets. 

In contrast to Gabriel, REVAMP\textsuperscript{2}T targets machine vision at the edge involving multiple cameras distributed across a geographical area. Unlike VisFlow, where all processing is done at the cloud, REVAMP\textsuperscript{2}T performs a considerable amount of the processing at the edge nodes (next to the camera) by custom-designed deep learning based vision engines, thereby decreasing bandwidth requirements. This also allows REVAMP\textsuperscript{2}T to protect the privacy of the tracked individuals, On the other hand, cloud-based systems must transfer personally identifiable information across large, interconnected networks and store that information in the cloud, where it is all too vulnerable. Additionally, the proposed edge vision system scales easily to a large number of cameras distributed over a wide geographic region.

REVAMP\textsuperscript{2}T accomplishes all these tasks online, in real-time, on low-power edge devices.


\vspace{-5pt}
\section{Privacy Requirements and Threat Modeling} \label{sec:Privacy}
\vspace{-5pt}

This section describes the privacy threat models which REVAMP\textsuperscript{2}T is designed to address. In safeguarding privacy, we wish to protect the identities and Personally Identifiable Information (PII) of the individuals being viewed by our system. This is most commonly in the form of raw image data, but can also refer to meta-data that can be used to determine the race, gender, nationality, or even identity of an individual. There are three main threats to this that we attempt to address:

\begin{itemize}
\item The external threat of someone getting unintended access to network communications and retrieving image data or Personally Identifiable Information (PII).
\item The internal threat of someone with authorized access to the system viewing image data or PII - even someone who is supposed to have access to the system should not be able to discern the identities of individuals or have access to their personal information. 
\item The physical threat of someone getting physical access to the edge device.
\end{itemize}

To safeguard against these threats, we impose two major policies for designing REVAMP\textsuperscript{2}T:

(1) REVAMP\textsuperscript{2}T will not store any image data or transfer it across the network. As soon as the image is processed on the edge node, it is destroyed. This protects any PII in the images from being viewed by anyone with access to the system. Even with direct access to the edge node, image data never touches non-volatile memory, so accessing it is impossible without fundamentally changing the semantics of our system.

(2) REVAMP\textsuperscript{2}T re-identification algorithm will work on an encoded feature representation of an individual (without using facial recognition algorithms). These features represent the visual and structural attributes of an individual, but can not be interpreted by humans and has zero meaning outside the constraints of our system. This means that even if a person gains access to this feature representation, intended or otherwise, they can not learn anything about the appearance or identity of the individual it was derived from. By utilizing these feature representations, REVAMP\textsuperscript{2}T is able to focus on \textit{differentiation} between people rather than personal \textit{identification}. This is in contrast to common methods that rely on facial recognition or other PII \cite{face-rgbd,facenet,face-gmm, face-eccv16, sighthound}.

We design REVAMP\textsuperscript{2}T, with respect to defined privacy protection policies. \secref{Algorithm} and \secref{System} present algorithmic constructs and IoT system design of REVAMP\textsuperscript{2}T.

\vspace{-5pt}
\section{REVAMP\textsuperscript{2}T:  Algorithmic Constructs} \label{Algorithm}
\vspace{-5pt}

\begin{figure}[t] 
	\centering
	\includegraphics[width=1\linewidth, trim= 2 2 2 2,clip]{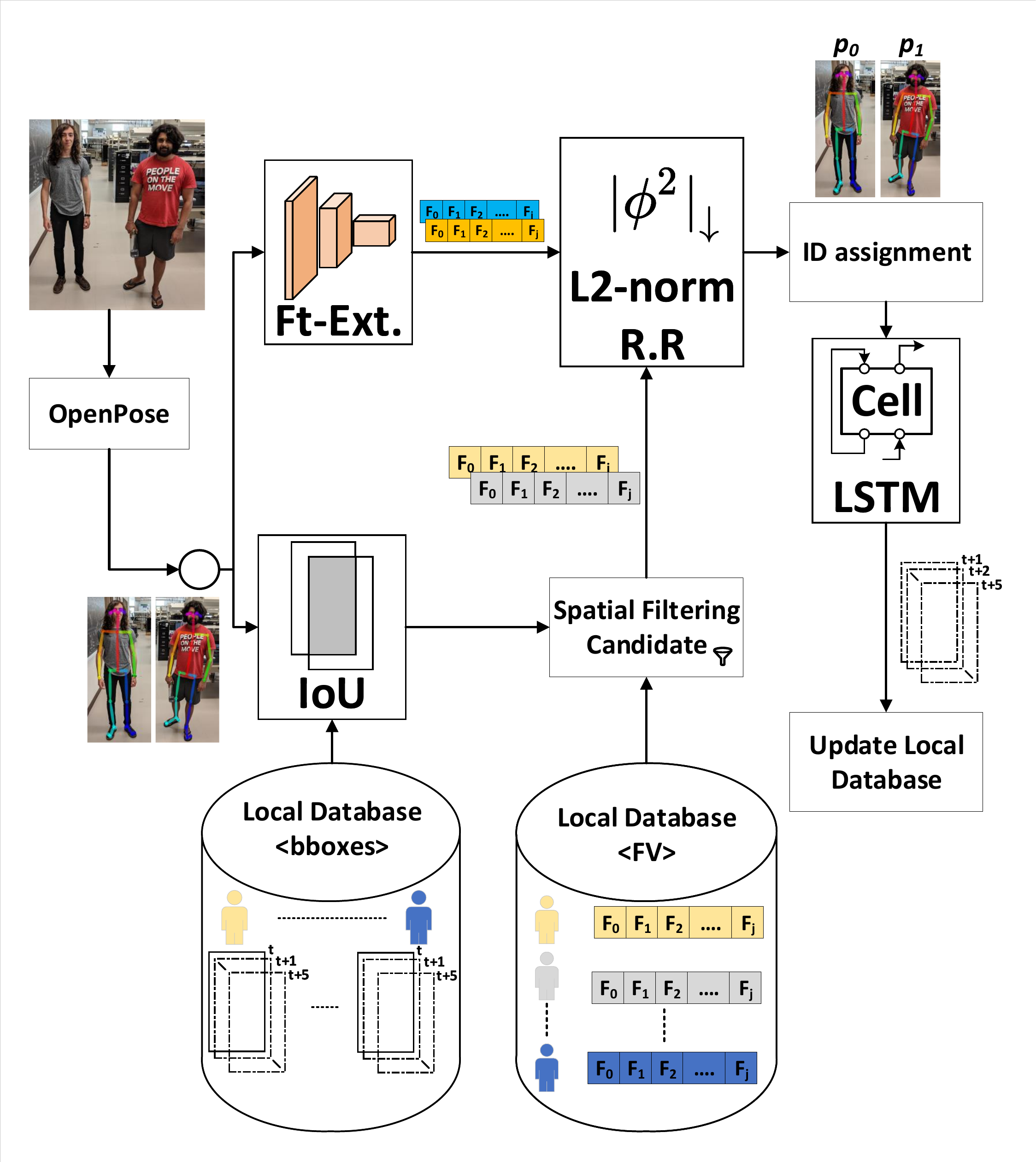}
	\caption{Algorithm Pipeline on the Edge}
	\label{fig:Algorithm_Pipeline}
	\vspace{-10pt}
\end{figure}

This section presents algorithmic constructs to enable real-time pedestrian re-identification and tracking while satisfying our privacy model detailed in \secref{sec:Privacy}.


\figref{fig:Algorithm_Pipeline} outlines the full algorithmic pipeline. The pipeline consists of three primary phases: (1) Detection, (2) Re-identification, and (3) Tracking and Prediction. For the detection part, we chose OpenPose \cite{openpose} from the CMU Perceptual Computing Lab. OpenPose is a pose prediction framework that uses part affinity fields to understand the image input and provide person detections with marked keypoint locations. In addition, it provides keypoints that reveal the motion of the human body, making them useful for motion prediction and action recognition. In the re-identification portion, discriminative features are generated for detection comparison and matching in the Local Database. Once the re-identification has been completed, an LSTM network is applied to predict future positions of known detections. The rest of this section discusses the technical details of our proposed re-identification, tracking, and integration.


\vspace{-15pt}
\subsection{Feature Extractor Network}\label{subsec:Re-Identification}
\vspace{-5pt}
The core of the re-identification is the feature extraction network to extract discriminative features from each detection, represented by the Ft-Ext. box in ~\figref{fig:Algorithm_Pipeline}. For this task, a deep convolution network had to be developed for accurate, real-time performance. Most deep convolution networks have a massive number of parameters and operations, which makes them computationally expensive for use in mobile and embedded platforms. MobileNet-V1 \cite{mobilenet} and \mn~\cite{mobileNetV2} are two developed light-weight deep convolution networks which effectively break down a standard convolution into a depth-wise and point-wise convolution to decrease the network parameters and operations. \mn~further improved the MobileNet-V1 architecture by adding linear bottleneck layers and inverted residual connections.

In this article, we use the \mn~model and change the fully connected layer to a 2D average pooling with the kernel size of (8, 4) in order to make the output of the network a 1x1280 vector as the embedded appearance features. We also use the triplet loss function \cite{Hermans2017InDO} to train the \mn~for extraction of discriminative features based on person appearance. The underlying architecture of a triplet loss network is consisted of three identical networks which transform the cropped RoI into embedding on a lower dimensional space. One RoI is the anchor image, the second is a positive sample of the anchor and third is a negative sample. The basic concept here is to minimize the distance between the anchor and the positive samples and maximize the distance between the anchor and the negative samples in the lower dimensional embedding space. To facilitate such learning, a suitable loss function is used after the embeddings are extracted from the RoIs:
\begin{equation}
Loss = \sum_{i=1}^{n}{\Big[ \alpha + \norm{f^{a}_{i}-f^{p}_{i}}^2 -\norm{f^{a}_{i}-f^{n}_{i}}^2 {\Big]}_{+} },
\end{equation}
where $\alpha$ is margin, $f^{a}$, $f^{p}$, and $f^{n}$ are embedded appearance features of the anchor, positive, and negative samples for the class $i$, respectively. Minimizing $Loss$ function will force all samples of class $i$ to be inside of hypersphere of radius $\alpha$. The dimension of the hypersphere is equal to the size of the network output (1280 for \mn). To further improve the performance of \mn~, we have assigned error-friendly operations, such as convolution and General Matrix Multiply (GeMM) operations, to half precision which is 16-bit floating point accuracy and applied mixed precision training \cite{mixedPrecision} to minimize the error caused by half precision operations.

\vspace{-5pt}
\subsection{Pedestrians Tracking}\label{subsec:Tracking}
After the re-identification process, we send the current detections to the LSTM network to get their respective bounding box predictions for the coming five frames. In this way, we can handle miss-detections that the detection network might suffer as it is running at the edge (lower detection network resolution). At the same time, we are able to handle short-term occlusions as we know the position of the occluded pedestrian and can ReID them once they are back in the scene.


To efficiently train our LSTM, we chose the DukeMTMC \cite{ristani2016MTMC} dataset for training the module as it focuses on pedestrian tracking. The dataset involves multiple targets, and we curate single target instances from the dataset so that we do not involve the re-identification pipeline for the training phase. This results in the network being trained on single instances and carrying out inference on multiple targets. Also, using this technique helps us to re-use the same model parameters for multiple pedestrians when predicting their future positions, saving us redundant computation and making our LSTM tracking module scalable. We leverage the sequence learning capabilities of LSTM by providing it with consecutive frame keypoints of a set sequence length and minimizing the mean-squared error between the obtained predictions and the ground-truth positions of the next five frames.

\begin{figure}[b] 
	\centering
	\vspace{-15pt}
	\includegraphics[width=1\linewidth,trim= 4 4 4 4,clip]{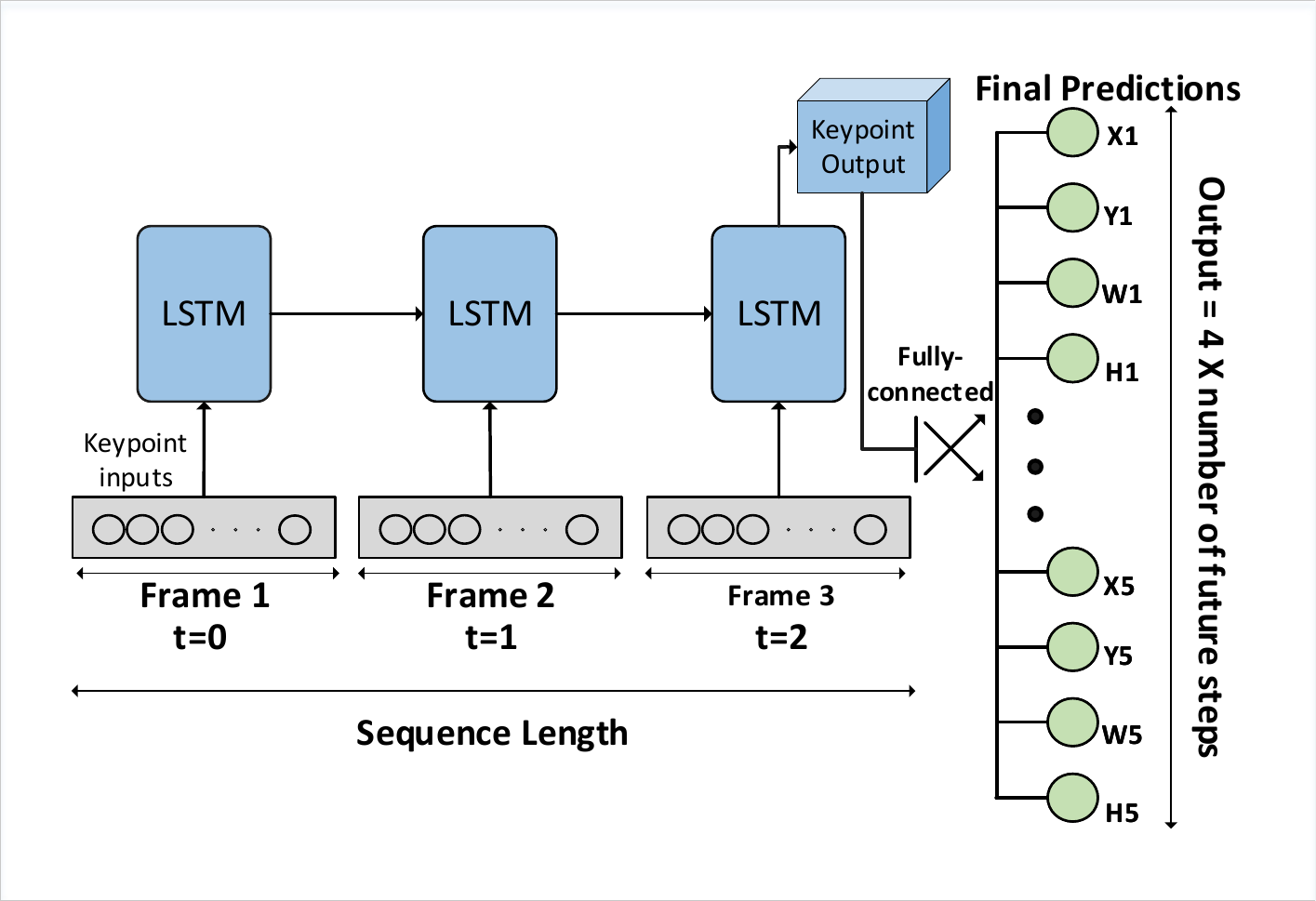}
		\vspace{-10pt}
	\caption{LSTM training: Feeding a sequence of frames and getting the predicted bounding box predictions}
	\label{fig:lstm}
\end{figure}

Fig~\ref{fig:lstm} shows our LSTM module in detail. We provide it with the keypoints of three (our sequence length) consecutive frames and send the last step output to the fully-connected layer which encodes these keypoints to the bounding box position of the pedestrian for the next five frames. The size of our trained LSTM model is under 1.5 MB, making it suitable to run on edge devices.

\subsection{Integration of Video Analytic Pipeline}
\label{subsec:integration}
In order for the entire re-identification task to be accomplished on the edge, these modules must be integrated seamlessly together. Referring back to \figref{fig:Algorithm_Pipeline}, a frame is inputted from the camera feed directly into the detection network. The resulting detections are received, scaled to the appropriate size and aspect ratio, and batched through the feature extractor (FT-Ext. box, \figref{fig:Algorithm_Pipeline}). The output of this network provides the encoded 1x1280 Feature Vector for each detection. In parallel, spatial filtering is being done on the Local Database (southern portion of \figref{fig:Algorithm_Pipeline}). For each detection, a subset of candidate matches are chosen based on IoU with last known or predicted bounding boxes from previous Local Database entries. The intuition behind this filtering mechanism is to ensure that detections are matched with entries that not only match in the embedded space (Feature Vector), but also in location and trajectory. Because the entire pipeline is running many times per second, the likelihood of a pedestrian traversing a substantial amount of distance or drastically changing trajectory between processed frames is low. Therefore, we avoid including entries that do not make sense from a positional standpoint in the candidate pool for ReID on a new detection.

Within the subset of candidates, the L2-norm operation can be done between Feature Vectors to differentiate between entries, and make final matching decisions using a re-ranking approach to ensure optimal ID assignment (L2-norm R.R box in ~\figref{fig:Algorithm_Pipeline}). As described above, the feature extraction network was trained to maximize the euclidean distance between Feature Vectors of different pedestrians, and minimize the distance between vectors of the same pedestrian. This training and inference methodology provides a privacy-aware approach to ReID, as per our threat models. Rather than using specialized, personally identifiable blocks of information to continually re-identify a pedestrian, our model simply encodes the current visual features of a detection to an abstract representation, and focuses on \textit{differentiation} between entries rather than \textit{personal} identification. Once all detections in the scene are assigned, the LSTM described previously takes in the detection keypoints and generates predicted bounding boxes for the next five time intervals. Finally, the Local Database is updated with assigned labels, keypoints, Feature Vectors, and generated predictions from the processed frame.

\section{REVAMP\textsuperscript{2}T: System Constructs}\label{System}
Creating algorithms that can effectively solve issues while running on low-power devices is of vital importance to enable inference on the edge. However, there are many system-level considerations that must be taken into account when developing a robust end-to-end system. How data flows between algorithms, when and how to utilize said algorithms, how to handle communications between the edge node and edge server, and how to map and optimize processes to and for the underlying hardware available on the edge. All of these are system-level design decisions that greatly impact the efficiency and viability of the end-to-end system. With REVAMP\textsuperscript{2}T's focus on privacy, it was important that the IoT system was designed around never storing any personally identifiable information. Algorithmic selections and the design of the system's processing flow hinged around that constraint.

\begin{figure}[h] 
	\centering
	\includegraphics[width=1\linewidth, trim= 2 2 2 2,clip]{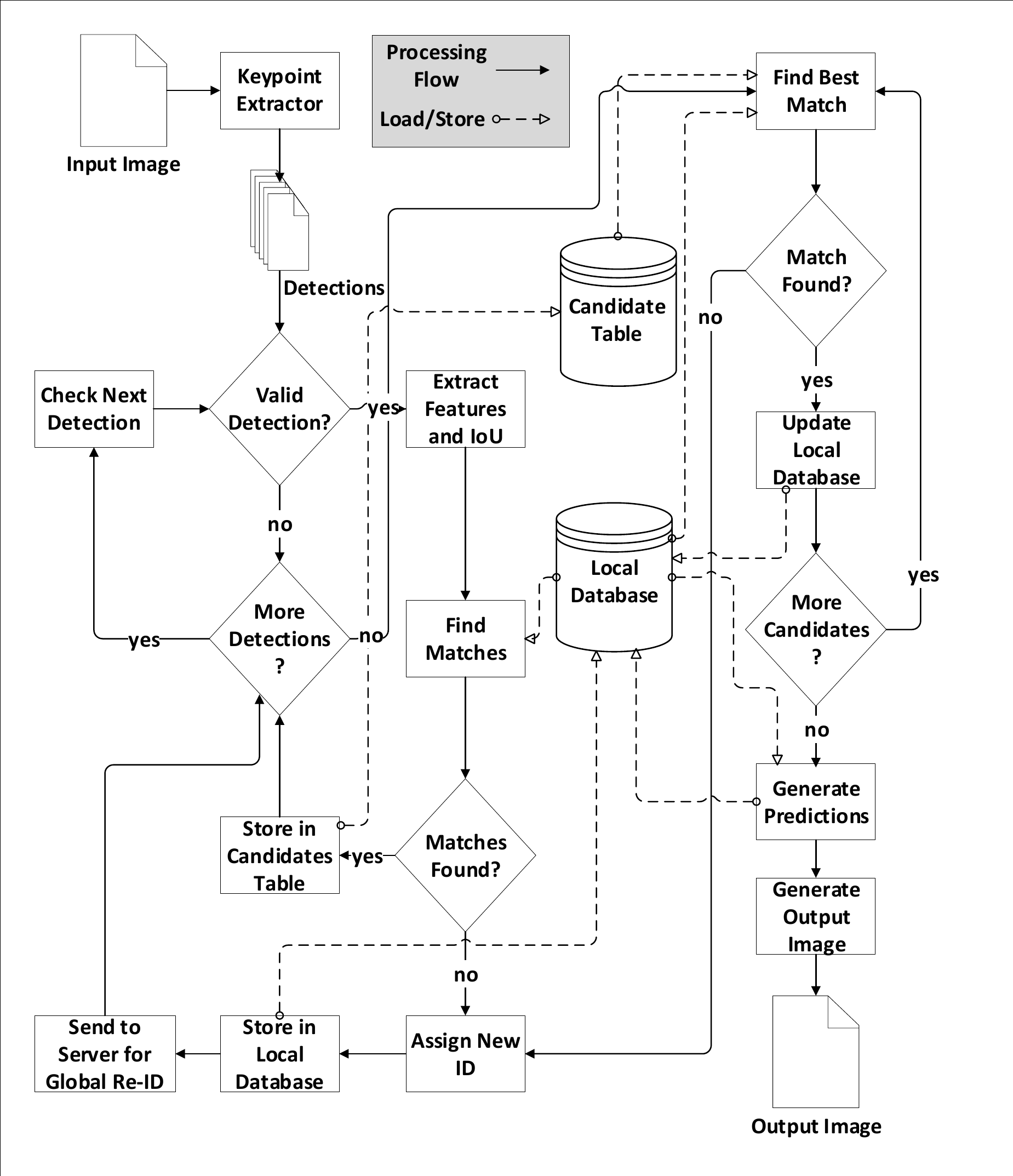}
	\captionsetup{justification=centering}
	\vspace{-20pt}
	\caption{Processing Flow of the Edge}
	\vspace{-20pt}
	\label{fig:ProcessingFlow}
\end{figure}

\vspace{-10pt}
\subsection{System Hyperparameters and Processing Flow}\label{sysflow}

\figref{fig:ProcessingFlow} shows the logical processing flow of one frame of data on the edge, beginning at when the image is extracted from the camera to when the final output is displayed on the edge device. First, the image is run through the keypoint extractor, which outputs a vector of detections. To remove false detections, each detected pedestrian should have a minimum number of keypoints equal to $\theta_{key}$, and each of those keypoints a confidence value of at least $\theta_{conf}$. 

\begin{table}[b]
    \centering
    \vspace{-10pt}
    \caption{System Parameters}
    \label{tab:hyperparameters}
    \begin{tabular}{|c|c|p{5cm}|}
        \hline
        \textbf{Parameter} & \textbf{Description} \\
        \hline
        $\theta_{key}$ & Minimum keypoints for valid detection \\
        $\theta_{conf}$ & Confidence threshold for valid keypoint \\
        $\theta_{euc}$ & Euclidean threshold with IoU \\
        $\beta_{euc}$ & Euclidean threshold without IoU \\
        $\theta_{IoU}$ & IoU threshold for updating Feature Vector \\
        $\Bar{D}$ & Detection from keypoint extractor \\
        $\Bar{V}$ & Valid Detections \\
        $\reallywidehat{DB}$ & Local database\\
        $\reallywidehat{C}$ & Candidate matrix\\
        $\zeta$ & Global variable to keep track of new ID\\
        \hline
    \end{tabular}
\end{table}

\vspace{-5pt}
\begin{figure*}[ht] 
	\centering
	\includegraphics[width=1\linewidth, trim= 2 2 2 15,clip]{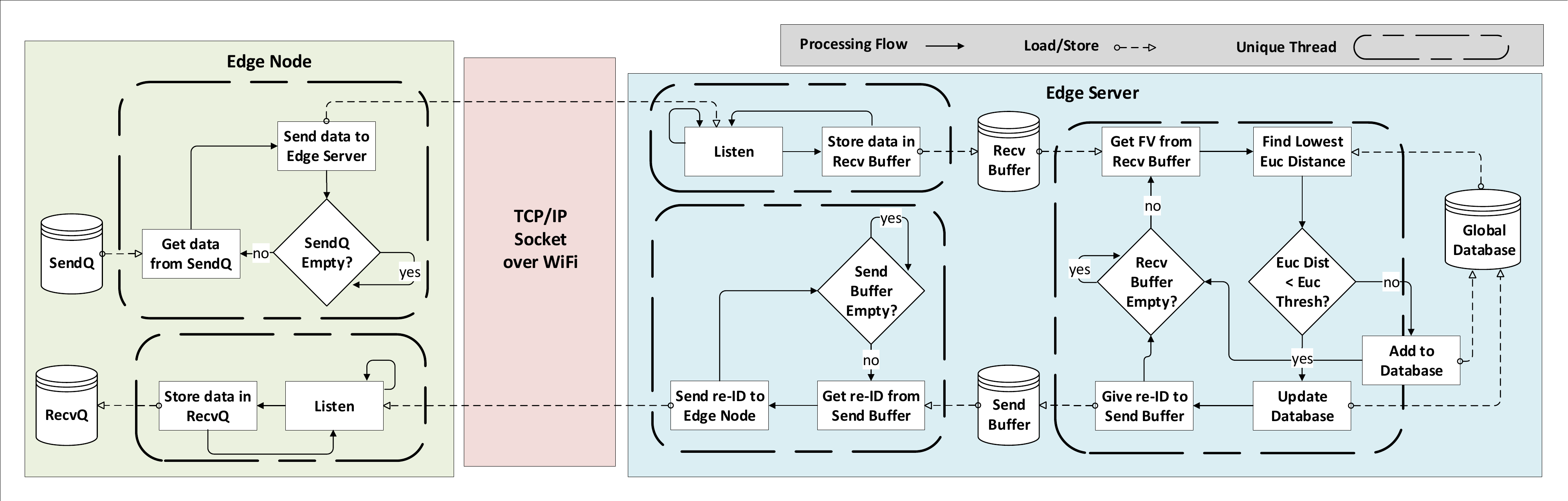}
	\captionsetup{justification=centering}
	\vspace{-20pt}
	\caption{Edge Node to Edge Server Communications}
	\vspace{-10pt}
	\label{fig:Comms}
\end{figure*}

\tabref{tab:hyperparameters} presents the system configurable hyperparameters. For every valid detection, all possible matches for ReID are gathered from the Local Database, as discussed in \secref{subsec:integration}. When a potential match is found, the detection, database entry index, and the Euclidean Distance between the two are stored in the Candidate Table. If no potential matches were found for a detection, it is considered to be a new person, assigned a new Local ID, stored in the Local Database, and sent to the server for Global ReID.



After all detections have been processed and the Candidate Table filled, the ReID processing is completed and IDs assigned, as shown in \algoref{alg:FindBestMatches}. The lowest Euclidean Distance score in the Candidate Table is found, the detection assigned the ID it was matched to, and the Local Database updated accordingly. Then all entries in the Candidate Table corresponding to that detection and Local Table entry are removed. This process is repeated until there are no suitable matches in the Candidate Table, after which all remaining detections are assigned new IDs.

For updating the Local Database on a successful ReID, the system always updates the spatial location of the person (bbox coordinates). However, it  only updates the Feature Vector if the IoU score is less than $\theta_{IoU}$ and the new Feature Vector is better representative of the object (meaning obtained with more keypoints than previously had). Whenever the Local Database is updated, a message is sent to the server to update its contents accordingly. Once ReID is complete, the system uses the LSTM to generate predictions on all applicable detections, as detailed in \secref{subsec:Tracking} and \secref{subsec:integration}.

\vspace{-20pt}
\subsection{Databases}
On the edge node, a "Local Database" is responsible for storing all pedestrians in the current scene. This database is filled with objects that contain IDs, bounding box coordinates, feature vectors, keypoints, and a parameter called life which keeps track of how many frames it has seen since that pedestrian has been detected. When an object has not been seen by the system after some time (as indicated by life), the object's ID is sent to the edge sever, informing the server of the object's removal from the Local Database. This has two main benefits. Reducing the length of time an object's data is stored on the edge increases the effectiveness of spatial reasoning through IoU, as well as ensuring any single person's data is not stored on the edge when they are not active in the current scene. It also acts as an efficient replacement policy without complex computation.

On the edge server, there is a "Global Database" that functions very similarly to the Local Database. It stores the exact same information, with the addition of knowing which edge node's scene, if any, the object is currently active in. When an object is active in an edge node's scene, that edge node gains ownership of that object, blocking it from being ReID'd by other edge nodes. This ownership is cleared when the server receives notification of the object's removal from the local database of the owning edge node, allowing the object to be included in ReID from all edge nodes. The size of both the Local Databases and the Global Database are easily configurable, allowing for customization to fit the requirements of individual applications.

\setlength{\textfloatsep}{0pt}
\begin{algorithm}[ht]
	\caption{Validating Detections}
	\label{alg:ValidDetection}
	\begin{algorithmic}[1]
		\Require $\Bar{D}, \theta_{conf}, \theta_{key}$
		\Ensure $\Bar{V}$
		\State $\Bar{V} \gets \emptyset$
		\For{$d$ in $\Bar{D}$}
		\State $numKeyPoints$ = findValidKeyPoints($d$, $\theta_{conf}$)
		\If{$numKeyPoints \geqslant \theta_{key}$}
		\State $\Bar{V} \gets \Bar{V} \cup \{d\}$
		\EndIf
		\EndFor
	\end{algorithmic}
\end{algorithm}

\setlength{\textfloatsep}{10pt}
\begin{algorithm}[ht]
	\caption{Finding Best Matches}
	\label{alg:FindBestMatches}
	\begin{algorithmic}[1]
		\Require $\reallywidehat{C}$
		\Ensure $\reallywidehat{newID}, \reallywidehat{reID}$
		\State $\reallywidehat{newID} \gets \Phi, \reallywidehat{reID} \gets \Phi$
		\State $\tilde{C} = $ sortBasedOnL2Norm($\reallywidehat{C}$)
		\For{$(v, e, \phi)$ in $\tilde{C}$}
		\If{$e \neq null$}
		\State $\reallywidehat{reID} \gets \reallywidehat{reID} \cup \{e.$id$\}$
		\State removeValidEntryAndCandidate($v$,$e$, $\tilde{C}$)
		\EndIf
		\EndFor
		\While{thereIsCandidate($\tilde{C}$)}
		\State $\zeta = \zeta + 1$
		\State $\reallywidehat{newID} \gets \reallywidehat{newID} \cup \{\zeta\}$
		\State removeValidEntryAndCandidate($v$,$\emptyset$, $\tilde{C}$)
		\EndWhile
	\end{algorithmic}
\end{algorithm}

\subsection{System Communication / Synchronization}
A vital aspect of REVAMP\textsuperscript{2}T's communication design hinges around exactly what data is sent to the server, according to our privacy threat models described in \secref{sec:Privacy}. The only data transmitted across the network is encoded Feature Vectors, impersonal IDs, and system metadata. The current iteration of REVAMP\textsuperscript{2}T's communication protocol leverages Wifi, but is adaptable and can be expanded to other communication protocols, such as LTE and 5G. This allows REVAMP\textsuperscript{2}T to keep up with the ever-changing communications landscape, no matter what technologies emerge.


\figref{fig:Comms} shows communication synchronization between edge nodes and edge server. Communications between the edge node and edge server are handled asynchronously. Both the edge node and edge server have separate threads to handle sending and receiving ReID information and storing it in separate buffers. These buffers hold the data until the main threads are ready to work on it. In addition to the main thread and the Global Database, the server has a separate set of these buffers and threads for each node. The main thread of the server processes this data in a round robin fashion and based on the metadata it receives from the edge node. It will either update the Global Database with a new Feature Vector for a global ID, release ownership of a global ID, or check the Global Database for ReID matches for the provided Feature Vector. Communications are only sent back to the edge node when a ReID match is successfully found. 

By handling all communications on separate threads unrelated to inference, communications are entirely decoupled from the processing pipeline, eliminating pipeline stalls that would normally result from inline communications. This decoupling also means that edge node throughput is not dependent on network latency; Local ReID will always perform at a constant FPS. Additionally, if the network goes down and communications are completely lost, the buffers will allow a level of data synchronization after communications are restored. The system efficiently utilizes edge resources, which has a monumental impact on inference time. On the server side, the proposed system achieves greater scalability, as edge nodes do not fight over available sockets and communications does not take away from ReID resources.

\begin{figure}[b] 
	\centering
		\vspace{-10pt}
	\includegraphics[width=0.9\linewidth, trim= 3 3 3 20,clip]{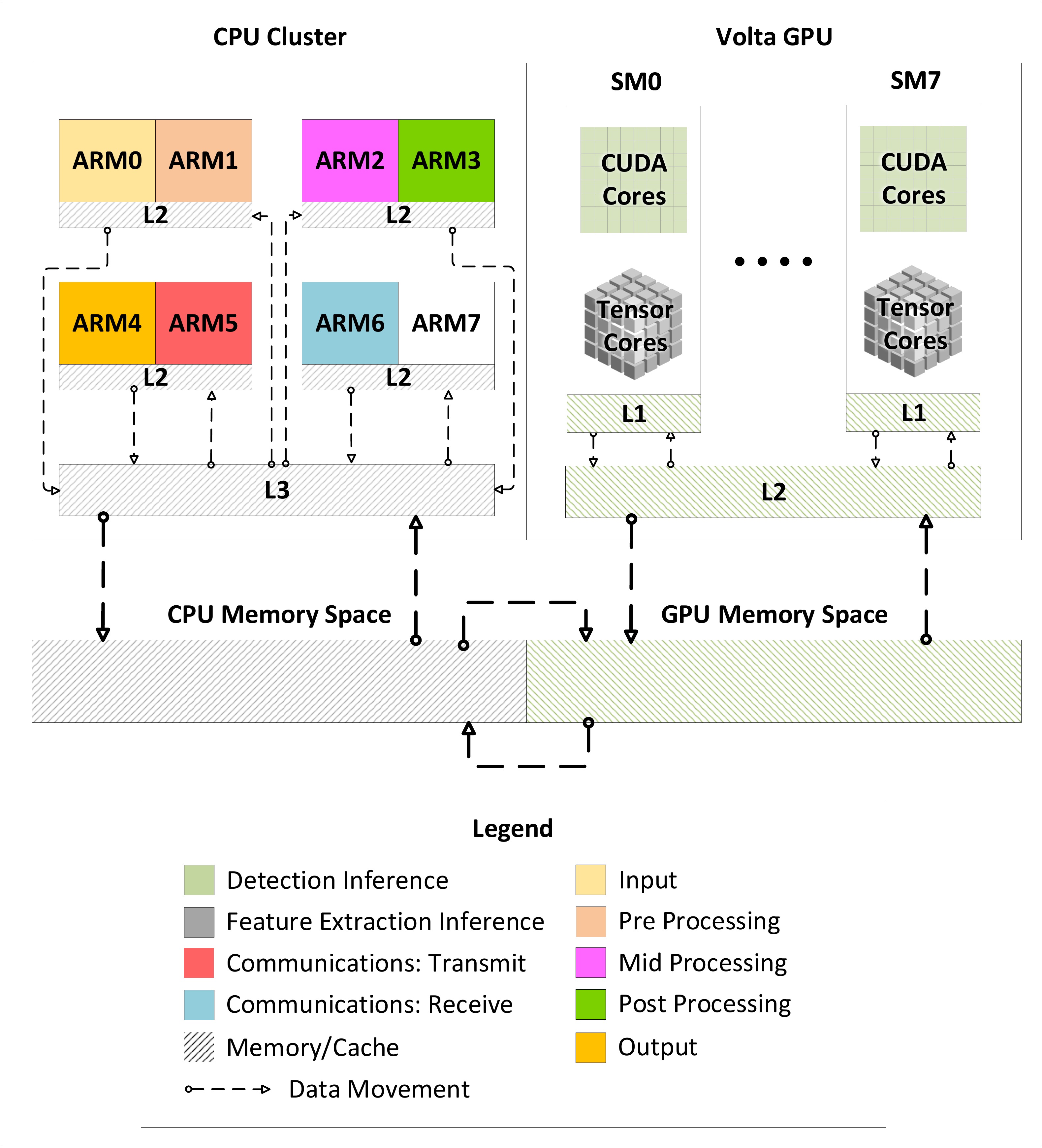}
	\captionsetup{justification=centering}
	\caption{Mapping of Processes to Edge Resources}
	\label{fig:EdgeMapping}
\end{figure}

\vspace{-20pt}
\subsection{Computation and Optimization}\label{Computation}
To achieve real-time performance on the edge, we chose Nvidia AGX Xavier SoCs \cite{Xavier}. The Xavier is equipped with many advanced components that are leveraged for REVAMP\textsuperscript{2}T, including eight ARM Core processors, two Nvidia Deep Learning Accelerators (NVDLA), and a Volta GPU with Tensor Cores optimized for FP16 Multiply and Accumulate. 

\figref{fig:EdgeMapping} shows the how the different processes in REVAMP\textsuperscript{2}T are mapped to the Xavier resources. Each stage of the detection framework is mapped to a separate ARM Core. The transmit and receive threads are mapped to their own cores as well. This leaves one ARM Core free to handle the OS and any background processes running outside of the system. Detection inference runs on the CUDA Cores of the Volta GPU. ReID inference is run on Tensor Cores. To enable this, the ReID network model was is converted from ONNX to use half precision through TensorRT. Batch normalization layers are also fused into the convolutional layers, reducing data migration. Detections are batched for ReID inference each frame, allowing a ReID throughput above 20 FPS. The NVDLAs were not used for ReID due to a lack of support for the level of group convolution in \mn. All code on the edge was developed in C++ for computational efficiency, enhanced execution, and mapping control.


\section{Experimental Results and Evaluation}\label{Results}
The experimental setups and results will be split into four subsections: Algorithm, System, Scalability, and Design Flexibility. All project code for simulations and full system implementation is provided on GitHub\footnote{\href{REVAMPT}{https://github.com/TeCSAR-UNCC/Edge-Video-Analytic}}.

\vspace{-10pt}
\subsection{Algorithm Evaluation}
\subsubsection{ Feature Extractor Network}
We used DukeMTMC-reID \cite{ristani2016MTMC,zheng2017unlabeled}, CUHK03 \cite{li2014deepreid}, and Market1501 \cite{zheng2015scalable} for evaluating the performance of two networks with different training methods. Table \ref{table:training_params} summarizes the hyperparameters of our network. We decreased learning rate exponentially after 150 epochs and used Adam optimizer to train both networks.

	

\begin{table}[h]
	\caption{Training Parameters}\label{table:training_params}
	\centering 
	\scalebox{1.0}{
		\begin{tabular}{|c|c||c|c|}
			\hline
			Description & Value & Description & Value
			\\[0.1ex]
			\hline                  
			Batch size  & 128 & Input shape (H$\times$W) & (256$\times$128)			
			\\
			IDs per batch  & 32 & Margin & 0.3	
			\\
			Instances per ID & 4 & Epoch & 300
			\\
			Initial learning rate   & $2\times10^{-4}$ & &
			\\
			\hline			
		\end{tabular}
	}
	
\end{table}

We used the baseline \rn~ as used in DeepCC for the sake of performance evaluation of \mn. We applied Cumulative Match Characteristic (CMC) \cite{moon2001computational, grother2003face} as a metric to evaluate and compare the identification performance of the two networks. Each dataset consists of a gallery  $\mathcal{G}$ as a set of various person images, and a query $\mathcal{Q}$ as a set of various person images that we want to identify. $\mathcal{P_G}$ is a $probe$ set, a subset of $\mathcal{Q}$, and for each of its images there are matches in $\mathcal{G}$. As the gallery embeddings are extracted, they are ranked (sorted) based on the similarity ($L2-Norm$ distance) across the current query image features. Then a set of the matched cases at rank $r$ can be defined as in \cite{grother2003face}:
\begin{equation}
\label{eq:ranked}
    C(r) = \{p_j~|~rank(p_j) \leqslant r\}~\forall{p_j} \in \mathcal{\mathcal{P_G}}
\end{equation}
Based on Eq.~\ref{eq:ranked} CMC at rank $r$ is calculated by following equation:
\begin{equation}
    CMC(r) = \frac{\left|C(r)\right|}{\left|\mathcal{\mathcal{P_G}}\right|} 
\end{equation}
It should be noted that the CMC calculation can still be different for each dataset. For example, in Market-1501, $\mathcal{Q}$ and $\mathcal{G}$ can share the same camera view. However, for each individual query image, the individual's samples in $\mathcal{G}$ from the same camera are excluded~\cite{zheng2015scalable}.

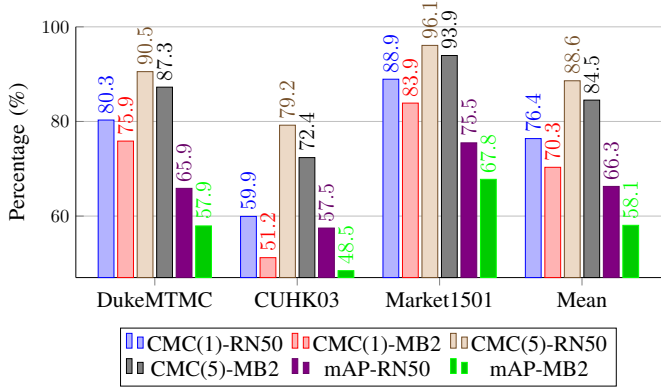
\begin{figure}[!tbp]
	\centering
		{\begin{tikzpicture}[scale=0.7]
		\begin{axis}[
		ybar,
		bar width=.3cm,
		width=0.70\textwidth,
		height=.35\textwidth,
		legend style={
			at={(0.5,-0.2)},
			anchor=north,
			legend columns=3,
			/tikz/every even column/.append style={column sep=0.1cm}
		},
		enlarge x limits={abs=1.5cm},
		x tick label style={font=\large, rotate=0, anchor=north},
		y tick label style={font=\normalsize},
		symbolic x coords={DukeMTMC,CUHK03,Market1501, Mean},
		xtick=data,
		nodes near coords = \rotatebox{90}{{\pgfmathprintnumber[fixed zerofill, precision=1]{\pgfplotspointmeta}}},
		nodes near coords align={vertical},
		ymajorgrids,
		ymin=47,ymax=100,
		ylabel={Percentage (\%)},style={font=\large},
        axis lines*=left,
        clip=false
    ]
		\addplot table[x=interval,y=cmc1-RN50]{\singleResNet};
		\addplot table[x=interval,y=cmc1-mb2]{\singleResNet};
		\addplot table[x=interval,y=cmc5-RN50]{\singleResNet};
		\addplot table[x=interval,y=cmc5-mb2]{\singleResNet};
		\addplot table[x=interval,y=mAP-RN50]{\singleResNet};
		\addplot table[x=interval,y=mAP-mb2]{\singleResNet};
		\legend{CMC(1)-RN50, CMC(1)-MB2, CMC(5)-RN50,CMC(5)-MB2, mAP-RN50, mAP-MB2}
		\end{axis}
		\end{tikzpicture}}
	\caption{ResNet-50 single precision and MobileNetV2 half precision accuracy evaluation on three different benchmarks} \label{fig:accResnetMb2}
\end{figure}

Another evaluation metric which gives a representation of network performance over a set of queries $\mathcal{Q}$ is mean Average Precision (mAP), which can be extracted by:
\begin{gather}
    mAP = \frac{\sum_{i=1}^{\left|\mathcal{Q}\right|}{AP(q_i)}}{\left|\mathcal{Q}\right|}, q_i \in \mathcal{Q}\\
    AP(q) = \frac{1}{TP_{gt}}\sum_{j}^{\left|\mathcal{G}\right|}{\frac{TP_{detected}}{j}}
\end{gather}
where $TP_{gt}$ is the number of ground-truth true positives, and $TP_{detected}$ is the number of true positives detected by the network. CMC(1), CMC(5), and mAP are computed and compared side-by-side in Fig.~\ref{fig:accResnetMb2} for both \rn~and \mn~ half precision networks. We can realize that the \mn~half precision is 6.1\% less than \rn~for mean value of CMC(1) across all three datasets, while reaching an 18.92$\times$ model size compression ratio for \mn~(5.0MB) over the baseline model (94.6MB).


\pgfplotstableread[row sep=\\,col sep=&, header=true]{
	interval      & cmc1-RN50   & cmc5-RN50  &  mAP-RN50 & cmc1-mb2   & cmc5-mb2  &  mAP-mb2\\
	DukeMTMC      & 80.30 & 90.53 & 65.86 & 75.85 & 87.25 & 57.92 \\
	CUHK03        & 59.93 & 79.21 & 57.47 & 51.21 & 72.36 & 48.46 \\
	Market1501    & 88.93 & 96.08 & 75.48 & 83.88 & 93.94 & 67.77 \\
	Mean          & 76.38 & 88.60 & 66.27 & 70.31 & 84.51 & 58.05 \\
}\singleResNet

\subsubsection{LSTM Prediction Network}
\begin{figure}[!tbp]
	\centering
\begin{tikzpicture}
\begin{axis}[x tick label style={
		/pgf/number format/1000 sep=0.1cm},
	ylabel=IoU,
	ymajorgrids,
	symbolic x coords={Cam1,Cam2,Cam3,Cam4,Cam5,Cam6,Cam7,Cam8},
	height=.25\textwidth,
    width=.5\textwidth,
	xtick=data,
	bar width=.5cm,
	x tick label style={font=\normalsize, rotate=25, anchor=north},
 	ybar, 
	ymin=0.3,ymax=0.65]
	
\addplot[fill=orange!60] table[x=interval,y=IoU]{\lstm};
\end{axis}
\end{tikzpicture}
	\caption{Average IoU for each camera on the testing sequences} \label{fig:lstmiou}
\end{figure}
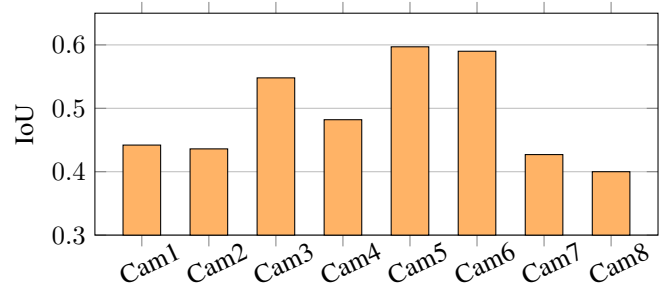

A total of 120 single object sequences, 15 from each camera, were used to train the LSTM model for 200 epochs. For testing the model we used 3 sequences per camera. The network takes around 14 hours to train on an Nvidia V100 GPU and was implemented using PyTorch. To evaluate the performance of the network we use the Intersection over Union (IoU) of the predicted bounding boxes with the ground truth bounding boxes and average it for all frames in the sequence, as shown in \figref{fig:lstmiou}. This average IoU shows that we maintain performance above the 0.3 IoU detection threshold typically used for evaluation. We do not compare these results with DeepCC because their approach uses tracklets rather than IoU for tracking evaluation.

\pgfplotstableread[row sep=\\,col sep=&, header=true]{
    interval    & IoU \\
    Cam1     & 0.442 \\
    Cam2    & 0.436  \\
    Cam3    & 0.548 \\
    Cam4   & 0.482 \\
    Cam5   & 0.597 \\
    Cam6      & 0.59 \\
    Cam7    & 0.427 \\
    Cam8   & 0.4 \\
    }\lstm

\subsubsection{Algorithm Pipeline}
In order to validate the accuracy of the full algorithm pipeline, the edge algorithms and edge server functionality were ported to MATLAB and compiled into a simulation testbed to gather results. For these experiments, we used the DukeMTMC dataset, which includes 85 minutes of 1080p footage from 8 different cameras on the Duke University campus. Specifically, the trainval\_mini frame set was used for validation. For comparison, we also ran the current state-of-the-art in MTMC work, DeepCC, on the same trainval\_mini validation set. For all experiments, we measure ID Precision (IDP), ID Recall (IDR), and ID F1 score (IDF1) with truth-to-result matching, as proposed in \cite{measures}. Intuitively, IDP measures the percentage of attempted ReIDs that were correct, and IDR measures the percentage of possible ReIDs completed, regardless of number of attempts. IDF1 is simply the harmonic mean of IDP and IDR. Detection misses are computed in accordance to the truth-to-matching method, with the IoU threshold at 0.3. In accordance with \tabref{tab:hyperparameters}, the values for system hyperparameters are as follows: $\theta_{key}$ = 5, $\theta_{conf}$ = 0.5, $\theta_{euc}$ = 5, $\theta_{euc\beta}$ = 2, $\theta_{IoU}$ = 0.3.

\begin{figure}[h]
	\centering
	\vspace{-5pt}
    \begin{tikzpicture} 
        \begin{axis}[ 
            legend entries={Multi, Cam1, Cam2, Cam3, Cam4, Cam5, Cam6, Cam7, Cam8},
            legend pos=outer north east,
            height=.32\textwidth,
            width=0.83\linewidth,
            xtick={0,10,...,100},
            ytick={0,10,...,100},
            xmin=40,
            xmax=100,
            ymin=70,
            ymax=100,
            xlabel=IDR,
            ylabel=IDP,
        clickable coords={(xy): \thisrow{label}}, 
        scatter/classes={
                dm={mark=square*,blue},
            d1={mark=triangle*,blue,fill=blue!50},
            d2={mark=oplus*,blue,fill=blue!50},
            d3={mark=diamond*,blue,fill=blue!50},
            d4={mark=star,blue,fill=blue!50},
            d5={mark=pentagon*,blue,fill=blue!50},
            d6={mark=10-pointed star,blue,fill=blue!50},
            d7={mark=halfcircle*,blue,fill=blue!50},
            d8={mark=halfsquare*,blue,fill=blue!50},
                rm={mark=square*,orange},
            r1={mark=triangle*,orange,fill=orange!50},
            r2={mark=oplus*,orange,fill=orange!50},
            r3={mark=diamond*,orange,fill=orange!50},
            r4={mark=star,orange,fill=orange!50},
            r5={mark=pentagon*,orange,fill=orange!50},
            r6={mark=10-pointed star,orange,fill=orange!50},
            r7={mark=halfcircle*,orange,fill=orange!50},
            r8={mark=halfsquare*,orange,fill=orange!50},
                rrm={mark=square*,black},
            rr1={mark=triangle*,black},
            rr2={mark=oplus*,black},
            rr3={mark=diamond*,black},
            rr4={mark=star,black},
            rr5={mark=pentagon*,black},
            rr6={mark=10-pointed star,black},
            rr7={mark=halfcircle*,black},
            rr8={mark=halfsquare*,black}}]
        \addlegendimage{only marks, mark=square*, black}
        \addlegendimage{only marks,mark=triangle*, black,fill=gray, mark options={scale=1.5}}
        \addlegendimage{only marks,mark=oplus*, black,fill=gray, mark options={scale=1.5}}
        \addlegendimage{only marks,mark=diamond*, black,fill=gray, mark options={scale=1.5}}
        \addlegendimage{only marks,mark=star ,black,fill=gray, mark options={scale=1.5}}
        \addlegendimage{only marks,mark=pentagon* ,black,fill=gray, mark options={scale=1.5}}
        \addlegendimage{only marks,mark=10-pointed star ,black,fill=gray, mark options={scale=1.5}}
        \addlegendimage{only marks,mark=halfcircle* ,black,fill=gray, mark options={scale=1.5}}
        \addlegendimage{only marks,mark=halfsquare*,black,fill=gray, mark options={scale=1.5}}
        \addplot[scatter,only marks,
            scatter src=explicit symbolic, mark options={scale=2}] 
        table[meta=label] { 
    x       y       label 
    79.00   79.24   dm 
    90.41   88.11   d1 
    93.78   91.17   d2
    89.88   90.27   d3 
    87.77   90.62   d4 
    85.02   83.45   d5 
    79.45   78.16   d6 
    89.49   84.86   d7 
    75.89   79.92   d8
      62.21    93.67   rm
    57.65    96.56    r1 
    67.62    94.01    r2
    74.96    96.07    r3 
    54.22    98.18    r4 
    79.05    96.74    r5 
    47.48    96.46    r6 
    82.70    97.94    r7 
    60.62    97.64    r8
        }; 

        \end{axis} 
    \end{tikzpicture}
    \caption{Precision (IDP) and Recall (IDR) for DeepCC (blue) and REVAMP\textsuperscript{2}T (orange)}  \label{fig:PvR}
    \vspace{-5pt}
\end{figure}
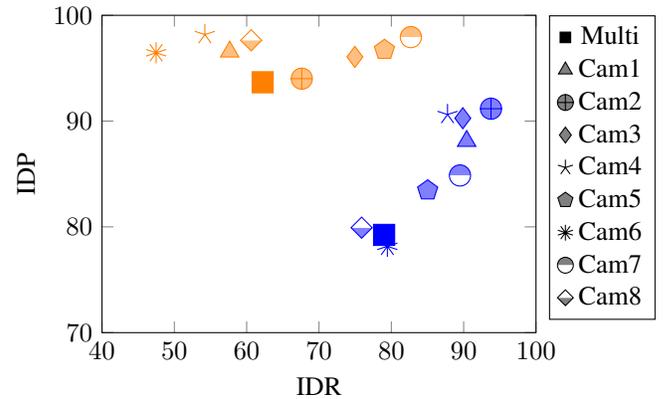

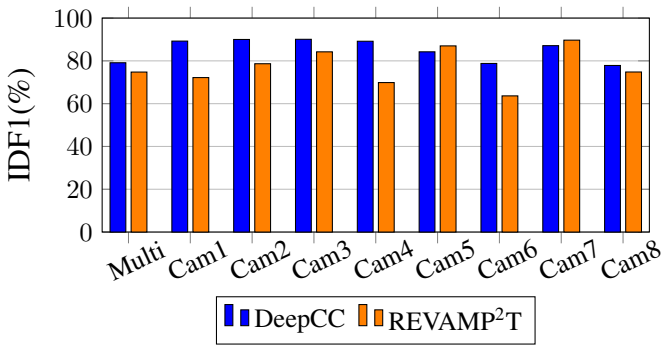
\begin{figure}[!tbp]
	\centering
		{\begin{tikzpicture}[scale=1.0]
		\begin{axis}[
		ybar,
		bar width=.21cm,
		width=1.0\linewidth,
		height=.5\linewidth,
		legend style={
			at={(0.5,-0.3)},
			font=\normalsize,
			anchor=north,
			legend columns=-1,
			/tikz/every even column/.append style={column sep=0.1cm}
		},
		enlarge x limits={abs=.35cm},
		x tick label style={font=\normalsize, rotate=25, anchor=north},
		y tick label style={font=\normalsize},
		symbolic x coords={Multi,Cam1, Cam2, Cam3, Cam4, Cam5, Cam6, Cam7, Cam8},
		xtick=data,
		ymajorgrids,
		nodes near coords align={vertical},
    	nodes near coords style={font=\large},
		ymin=0,ymax=100,
		ylabel={IDF1(\%)},style={font=\large},
		]
		\addplot[fill=blue] table[x=interval,y=dcc]{\FvCam};
		\addplot[fill=orange] table[x=interval,y=r5]{\FvCam};
		\legend{DeepCC, REVAMP\textsuperscript{2}T}
		\end{axis}
		\end{tikzpicture}}
	\caption{IDF1 Results for Multi-Camera and Single Camera}
	\vspace{-10pt}
	\label{fig:FvCameras}
\end{figure}

\figref{fig:PvR} shows IDP versus IDR for REVAMP\textsuperscript{2}T and DeepCC. Analyzing the results, DeepCC maintains groupings around 80\% for both IDP and IDR. REVAMP\textsuperscript{2}T maintains high IDP, always above 90\%; however, the IDR is less consistent across cameras. The reasons for this problem are two fold. First, because REVAMP\textsuperscript{2}T is an online system, it was designed to ReID within a short temporal window, in accordance with the spontaneous nature of online operation. Second, many of our false negatives are simply the result of missed detections. For the full 8-camera (shown as Multi) scenario, 59\% of the false negatives incurred were from missed detections from the first stage of the pipeline. As mentioned in \secref{Algorithm}, we chose to run the detection network at a relatively low resolution at an attempt to balance reasonable runtime speed and detection accuracy. Nonetheless, despite the challenges of edge-capable algorithmic development, REVAMP\textsuperscript{2}T maintains reasonably close IDF1 in comparison to DeepCC. \figref{fig:FvCameras} shows the IDF1 for each camera individually, as well as the complete 8-camera global system (Multi) for both approaches. Overall, REVAMP\textsuperscript{2}T only drops 4.3\% IDF1 in the full multi-camera system compared to the offline DeepCC algorithm, with DeepCC at 79.1\% and REVAMP\textsuperscript{2}T at 74.8\%.

\pgfplotstableread[row sep=\\,col sep=&, header=true]{
	interval      & dcc   & r5\\
	Multi        & 79.12 & 74.77\\ 
	Cam1      & 89.24 & 72.19\\
	Cam2      & 90.00 & 78.66\\
	Cam3      & 90.08 & 84.22\\
	Cam4      & 89.17 & 69.86\\
	Cam5      & 84.23 & 87.00\\
	Cam6      & 78.80 & 63.64\\
	Cam7      & 87.11 & 89.68\\
    Cam8      & 77.85 & 74.80\\
}\FvCam

\vspace{-0.55cm}
\subsection{System Evaluation}
For all measurements, REVAMP\textsuperscript{2}T is run in real-time. We also compare against DeepCC \cite{DeepCC}. For both, 16 detections per frame is assumed. As DeepCC was not built as a real-time system, it would be unfair to include the latency incurred through gallery matching in these comparisons, so we ignore the effect of this on power and latency. Real-time candidate matching is built into REVAMP\textsuperscript{2}T, so it is included in all reported measurements. For measuring the power consumption on the \nx, the Tegrastats was used. 

\subsubsection{Power Consumption and Computation Efficiency}
For power consumption on the Titan V and V100 GPUs, we utilized the NVIDIA System Management Interface. AMD $\mu$Prof was used to measure CPU idle power for the edge server. For REVAMP\textsuperscript{2}T, 1080p 30 FPS video was pulled directly from a webcam. For DeepCC, 1080p 60 FPS video was read from memory. In both cases, a brief warm up of 20 frames was allowed before power was sampled over 100 frames. Measurements for FPS were taken directly from the OpenPose GUI.

\begin{table}[h]
    \centering
    \caption{FPS and Power Consump. of Real-Time Inference}
    \label{tab:PowerAndFPS}
    \begin{tabular}{|C{1cm}|C{1.4cm}|C{1.2cm}|C{1.2cm}|C{1.2cm}|}
        \hline
        \textbf{System} & REVAMP\textsuperscript{2}T & DeepCC & DeepCC & DeepCC  \\
        \hline
        \textbf{Device} & Xavier & Titan V & 2xTitan V & V100 \\
        \textbf{FPS $\uparrow$} & 5.7 & 2.5 & 4.7 & 2.7 \\
        \textbf{Power $\downarrow$} & 34.4W & 200W & 365W & 224W \\
        \hline
        \hline
        \multicolumn{5}{|c|}{\textbf{Detailed Xavier Power Consumption}} \\
        \hline
        CPU & GPU & DDR & SOC & Total \\
        \hline
        4.2W & 22.1W & 2.75W & 5.35W & 34.4W \\
        \hline
    \end{tabular}
\end{table}

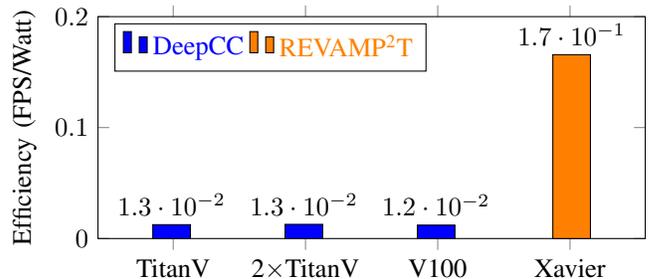
\begin{figure}[b]
    \centering{
\begin{tikzpicture}
\begin{axis}[
    width=1.0\linewidth,
	height=.25\textwidth,
    legend columns=2,
    legend pos=north west,
    ybar,
    enlarge x limits={abs=1.0cm},
    bar width=0.5cm,
    ymin=0,
    ymax=.2,
    xmin=0,
    xmax=3,
    ytick={0,0.1,...,0.3},
    ylabel={Efficiency (FPS/Watt)},
    		nodes near coords = \rotatebox{0}{{\pgfmathprintnumber[fixed zerofill, precision=1]{\pgfplotspointmeta}}},
		nodes near coords align={vertical},
    nodes near coords align={vertical},
    xtick={0,1,2,3},
    xticklabels={ TitanV , 2$\times$TitanV , V100 , Xavier },
    x tick label style={rotate=0,anchor=north,xshift=0cm}
    /tikz/every even column/.append style={column sep=1cm}
]

\addplot [fill=blue] 
    coordinates {(0.15, .0125) (1.15, .0128) (2.15, .0121)};
\addlegendentry[color=blue]{DeepCC}

\addplot[fill=orange] 
    coordinates {(2.85, .1657)};
\addlegendentry[color=orange]{REVAMP\textsuperscript{2}T}

\end{axis}
\end{tikzpicture}}
\caption{Efficiency of each test case.}
    \label{fig:Efficiency}
\end{figure}

\tabref{tab:PowerAndFPS} presents the power consumption and FPS for REVAMP\textsuperscript{2}T and DeepCC. Here we can see that for real-time applications, REVAMP\textsuperscript{2}T out performs DeepCC on each GPU setup we tested. Even using two Titan V's, DeepCC is only able to reach 4.7 FPS. Meanwhile, REVAMP\textsuperscript{2}T can reach 5.7 FPS. In addition, REVAMP\textsuperscript{2}T consumes only 17\% of the power of DeepCC on a single Titan V, or 9\% for the dual Titan setup. \figref{fig:Efficiency} presents computation efficiency, which is FPS processing per watt. DeepCC has an Efficiency between 0.0147 and 0.0161 FPS/Watt in all configurations. In comparison, REVAMP\textsuperscript{2}T has an Efficiency of 0.166 FPS/Watt. When looking at Efficiency, REVAMP\textsuperscript{2}T performs an order of magnitude better than DeepCC for real-time applications. This is because REVAMP\textsuperscript{2}T was built from the ground up to perform in real-time, both algorithmically and systemically.

\subsubsection{Accuracy $\bullet$ Efficiency (\AE)}
To enable real-time AI applications on the edge, we propose a new metric with which to measure edge performance; that is Accuracy $\bigcdot$ Efficiency (\AE). With \AE, we combine the algorithmic measurement of Accuracy with the systemic measurement of Efficiency to measure how well an application will perform in a real-time edge environment. \AE~has two parts: an \AE~mark, which is a score measured by the product of Accuracy and Efficiency, and \AE~coverage, which is measured in area, as determined by all the components of an \AE~mark. The components in \AE~coverage, when not already reported as a percentage, are normalize to be so. In the case of power, this normalized value is subtracted from one, as lower power consumption is preferable.

\begin{figure}[h]
	\centering
    \begin{tikzpicture} 
        \begin{axis}[ 
            legend entries={DeepCC, REVAMP\textsuperscript{2}T},
            legend style={at={(0.28,0.2)},anchor=west},
            height=.25\textwidth,
            width=1.0\linewidth,
            xtick={0,0.05,...,0.3},
            ytick={0,20,...,100},
            xmin=0,
            xmax=.2,
            ymin=60,
            ymax=85,
            xlabel=Efficiency (FPS/Watt),
            ylabel=Accuracy (IDF1\%),
        clickable coords={(xy): \thisrow{label}}, 
        scatter/classes={
            dm={mark=oplus*,blue,fill=blue},
            d3={mark=diamond*,orange}
        }]
        \addlegendimage{only marks,mark=oplus*,blue,fill=blue}
        \addlegendimage{only marks,mark=diamond*,orange,mark options={scale=1.5}}
        \addplot[scatter,only marks,
            scatter src=explicit symbolic, mark options={scale=2}] 
        table[meta=label] { 
    x       y       label 
    .01288  79.116   dm 
    .1657   74.77    d3 
        }; 
        \end{axis} 
    \end{tikzpicture}
    \caption{\AE~of DeepCC on Titan V and REVAMP\textsuperscript{2}T on Xavier.}  
    
    \label{fig:AE}
\end{figure}
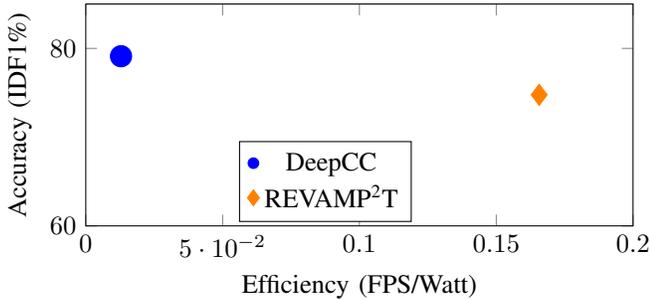

\begin{table*}[h]
    \small
    \centering
    \caption{Scalability Evaluation Results}
    \label{tab:scalability}
    \resizebox{.99\linewidth}{!}{
    \begin{tabular}{|c|c|c|C{1cm}|C{1.7cm}|C{1.7cm}|c|C{1cm}|C{1.7cm}|C{1.7cm}|c|C{1cm}|C{1.7cm}|C{1.7cm}|}
        \hline
        & \multicolumn{5}{c|}{\textbf{Server Processing}} & \multicolumn{4}{c|}{\textbf{Split Processing}} & \multicolumn{4}{c|}{\textbf{Edge Processing}} \\
        \hline
        \textbf{Nodes} & GPUs & Cost & Power (W) & End-to-end Latency (ms) & Network Latency (ms) & Cost & Power (W) & End-to-end Latency (ms) & Network Latency (ms) &  Cost & Power (W) & End-to-end Latency (ms) & Network Latency (ms) \\
        \hline
        1  & 1  & \$8,300  & 201  & \textbf{\blue{83.2}}   & 21   & \$8,900  & 122  & 506.6  & 29   & \textbf{\textcolor{ao}{\$5,700}}  & \textbf{\textcolor{amber}{34}}   & 540.6                  & 17  \\
        2  & 1  & \$8,400  & 202  & \textbf{\blue{105.0}}  & 35   & \$9,600  & 203  & 571.6  & 53   & \textbf{\textcolor{ao}{\$6,400}}  & \textbf{\textcolor{amber}{69}}   & 549.6                  & 26  \\
        4  & 2  & \$11,800 & 359  & \textbf{\blue{168.8}}  & 51   & \$11,000 & 306  & 678.6  & 82   & \textbf{\textcolor{ao}{\$7,800}}  & \textbf{\textcolor{amber}{138}}  & 559.7                  & 36  \\
        8  & 3  & \$15,400 & 518  & \textbf{\blue{223.8}}  & 107  & \$13,800 & 455  & 941.6  & 175  & \textbf{\textcolor{ao}{\$10,600}} & \textbf{\textcolor{amber}{275}}  & 601.9                  & 78  \\
        16 & 6  & \$25,000 & 946  & \textbf{\blue{374.8}}  & 258  & \$19,400 & 742  & 1547.1 & 414  & \textbf{\textcolor{ao}{\$16,200}} & \textbf{\textcolor{amber}{550}}  & 700.3                  & 175 \\
        32 & 11 & \$48,400 & 1827 & 972.8                  & 856  & \$30,600 & 1301 & 2723.2 & 858  & \textbf{\textcolor{ao}{\$27,400}} & \textbf{\textcolor{amber}{1101}} & \textbf{\blue{866.0}}  & 338 \\
        64 & 22 & \$91,800 & 3609 & 1202.8                 & 1086 & \$53,000 & 2406 & 6020.9 & 2074 & \textbf{\textcolor{ao}{\$49,800}} & \textbf{\textcolor{amber}{2202}} & \textbf{\blue{1095.7}} & 559 \\
        \hline
    \end{tabular}
    \vspace{-25pt}
    }
\end{table*}

\begin{figure}[h] 
	\centering
		\vspace{-10pt}
	\includegraphics[width=0.9\linewidth, trim= 50 180 60 190  ,clip]{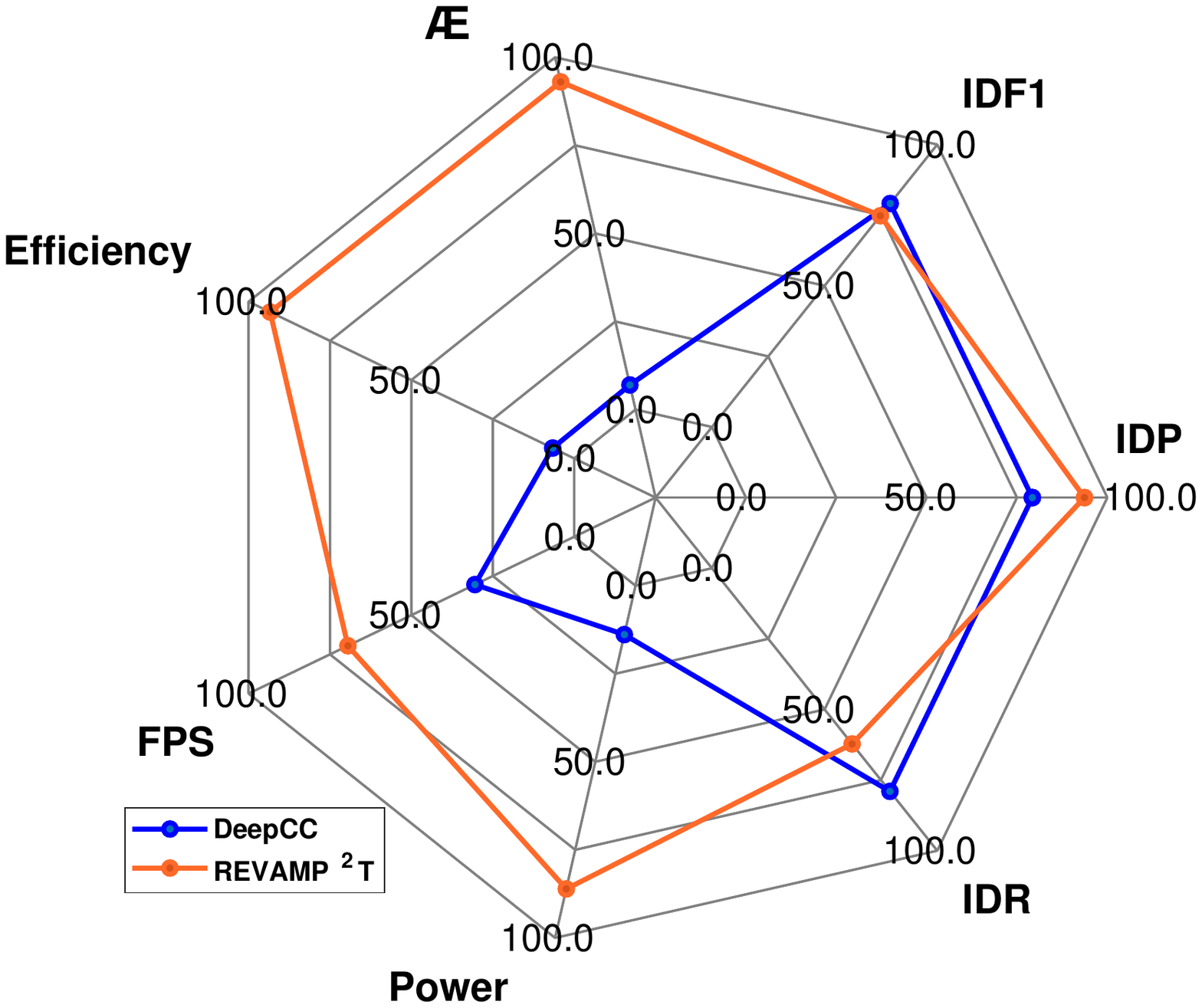}
	\captionsetup{justification=centering}
	\vspace{-10pt}
	\caption{\AE~Coverage}
	\label{AECoverage}
	\vspace{-10pt}
	\label{fig:AECoverage}
\end{figure}

In the case of REVAMP\textsuperscript{2}T, Accuracy would take the form of F1, while Efficiency is measured in FPS/Watt. \figref{fig:AE} shows the \AE~Mark for REVAMP\textsuperscript{2}T and DeepCC, while \AE~Coverage can be seen in \figref{AECoverage}. Here you can see that while DeepCC outperforms REVAMP\textsuperscript{2}T in terms of IDR and IDF1 Accuracy, REVAMP\textsuperscript{2}T has a significantly higher \AE~Mark (12.39 vs 1.02) and almost double the total \AE~Coverage (81.25\% vs 42.75\%). This is because our optimizations allow us to operate at twice the framerate, 17\% of the power, and we only lose by 4.3\% in F1 accuracy.

\subsection{Scalability}
To measure scalability, we compare REVAMP\textsuperscript{2}T, which pushes all local processing to the edge nodes, against two other configurations: (1) "Server Processing" which streams the edge video frames to the edge server to handle all the local processing, and (2) "Split Processing" which splits local processing between the edge node and the edge server. \tabref{tab:scalability} lists the number of GPUs, cost, power, end-to-end latency, and network latency over increasing number of edge nodes for all three scenarios. Power was measured using the same methods as described previously. The latency for a single frame was measured from when it was grabbed from the camera/video, to right before being displayed (using chrono library in C++), averaged over 100 frames, after a 20 frame warm up. 
For Server Processing, the edge nodes are Nvidia Jetson Nano SoCs \cite{nano}, as they only stream video to the server. For the other cases, the Nvidia Xavier was used. In all cases we assume an edge server with a 12 core CPU, at least 32 GB of memory, and the capacity to support up to eight Nvidia Titan-V GPUs. The server processes all data at 30 FPS. We assume video data from separate nodes can be interwoven to allow a single instance of REVAMP\textsuperscript{2}T to support two edge nodes. Network latency was simulated using NS-3 Discrete Network Simulator, using 802.11ac, TCP, and 600Gbps throughput. H.264 compression for video, PNG for images, and 16 detections per frame are assumed.






From \tabref{tab:scalability} we are able to see that Server Processing is able to achieve the lowest latency for smaller node counts. Past 16 edge nodes, the network latency from streaming video tips the scales latency to favor Edge Processing. However, Edge Processing always wins out in terms of cost and power consumption, making it a promising option particularly for high node counts. At 64 edge nodes, Edge Processing has the best latency while only requiring about half the cost and 60\% the power of Server Processing. We expect this trend to continue on to even higher node counts, increasingly favoring Edge Processing. Split Processing fares the worst in the comparison, due mostly to network latency. PNG compression is not nearly as efficient as H.264, meaning far more data is sent across the network, leading to the large latencies seen in the table. Overall, this shows that even for computationally intensive applications, edge processing is the only truly scalable solution for real-time IoT applications.

\begin{table}[H]
    \caption{Design Configuration Analysis}
    \label{tab:DCA}
\centering
\begin{tabular}{ |C{0.9cm}|C{1.5cm}|C{0.8cm}|C{1.4cm}| } 
\hline
\textbf{Config.} & \textbf{Power(W) $\downarrow$} & \textbf{FPS $\uparrow$} & \textbf{Accuracy $\uparrow$}\\ 
\hline
P\textsubscript{2}  & \textbf{\textcolor{amber}{12.48}} & 3 & 71.37\% \\ 
\hline
P\textsubscript{3} & 15.51 & 4 & 73.67\% \\ 
\hline
C\textsubscript{D} & 34.40 & 5 & 74.80\% \\
\hline
R\textsubscript{720} & 36.47 & 3 & \textbf{\blue{74.91\%}} \\
\hline
R\textsubscript{256} & 30.09 & 15 & 73.77\% \\
\hline
R\textsubscript{128} & 26.01 & \textbf{\textcolor{ao}{24}} & 0.97\% \\
\hline
\end{tabular}
\vspace{-5mm}
\end{table}

\subsection{Design Flexibility and Adaptation}

REVAMP\textsuperscript{2}T can further be configured to prioritize accuracy, FPS, or power consumption, as illustrated \tabref{tab:DCA}. The default configuration of REVAMP\textsuperscript{2}T is shown as C\textsubscript{D}, with an input resolution of 496x368 for the keypoint extractor, and power consumption as seen in \tabref{tab:PowerAndFPS}. We analyze five additional design configurations by modifying the input resolution, as well as the power restriction levels on the Xavier device. Configuration R\textsubscript{720}, R\textsubscript{256}, and R\textsubscript{128} are the proposed REVAMP\textsuperscript{2}T with modified input resolutions at 720x544, 256x192, and 128x96. Configuration P\textsubscript{2} and P\textsubscript{3} are the proposed REVAMP\textsuperscript{2}T configured with Power Mode 2 and 3 (C\textsubscript{D} uses Power Mode 0) provided by the Xavier device \cite{power_mode}.

 While R\textsubscript{720} does provide slightly higher accuracy than other configurations, it does incurs loss in FPS and increased power consumption. The keypoint extraction resolution for R\textsubscript{128} cannot properly extract keypoints for persons further than $\sim$30 feet from the camera, resulting in low accuracy for DukeMTMC. R\textsubscript{256} offers an option for additional throughput at an accuracy loss. While R\textsubscript{256} performs well on the tested dataset, we found that in real-world testing, this configuration was only able to extract keypoints for persons within $\sim$50 feet from the camera. Therefore, for robustness to real-world situations and its balance across all areas, configuration C\textsubscript{D} was chosen for the analysis of this report. With deployment in an IoT environment, C\textsubscript{D} would likely require PoE Type 3. The P\textsubscript{3} and P\textsubscript{2} configurations show how REVAMP\textsuperscript{2}T could be adapted to the power levels of PoE Type 2 and Type 1 for deployment, with minimal loss in accuracy.
 

\vspace{-10pt}
\section{Conclusions} 
\label{Conclusion}
This article proposed REVAMP\textsuperscript{2}T as an integrated end-to-end IoT system to enable decentralized edge cognitive intelligence for situational awareness. For the results and evaluation, this article also proposed a new two-part metric, Accuracy$\bigcdot$Efficiency (\AE). REVAMP\textsuperscript{2}T outperforms current state-of-the-art by as much as a thirteen-fold improvement in \AE. Future directions include designing light-weight human feature extractors, as a replacement for OpenPose,  further improvement of edge performance by designing application-specific hardware on FPGAs, data encryption to secure communications,  and containerizing REVAMP\textsuperscript{2}T using edge Kubernetes for scalable system orchestration and remote programmability across the edge devices.



\vspace{-10pt}
\section*{Acknowledgement}
\vspace{-5pt}
This research is supported by the National Science Foundation (NSF) under Awards No. 1737586 and 1831795.
\vspace{-5pt}


\bibliographystyle{IEEEtran}
\scriptsize
\bibliography{references}

\end{document}